# 烟花算法研究综述

赵志刚[1]，李智梅*，莫海淼，曾敏

广西大学，计算机与电子信息学院，南宁，530004

**摘 要**：烟花算法是一种新型智能优化算法，由于其收敛速度快，易于实现，并且还具有爆发性、多样性、简单性和随机性等特点，近来在许多研究领域受到越来越多的关注。本文介绍了烟花算法的提出背景、组成、改进思想（对算子的分析与改进、混合算法的改进），以及该算法在连续优化问题、离散优化问题、单目标优化问题、多目标优化问题等领域的相关应用。最后总结了烟花算法未来在理论分析、算子分析与改进、混合算法研究、算法应用这几个方面的研究方向。

**关键词**：群体智能；烟花算法；爆炸算子；变异算子；研究综述

中图分类号： TP301.6　　　文献标志码： A

## Review of research on fireworks algorithm

Zhao Zhigang, Li Zhimei*, Mo Haimiao, Zeng Min

*School of Computer and Electronic Information, Guangxi University, Nanning 530004，China*

**Abstract**：Fireworks algorithm is a new type of intelligent optimization algorithm. Because of its fast convergence speed, easy implementation, explosiveness, diversity, simplicity and randomness, it has attracted more and more attention in many research fields recently. This paper introduces the background, composition, improvement idea of fireworks algorithm (analysis and improvement of operator, improvement of hybrid algorithm), and its application in continuous optimization, discrete optimization, single-objective optimization, multi-objective optimization and other fields. Finally, the future research directions of fireworks algorithm are summarized, including theoretical analysis, operator analysis and improvement, hybrid algorithm research and algorithm application.

## 1. 概述

由于计算复杂理论的发展以及工业应用的迫切需要，不少科研工作者投入到群体智能算法的研究中，以便为人类更好地服务。群智能算法[1]是指在简单个体之间通过互相协作能够解决复杂问题，而这种能力是简单个体不具备的。

根据研究对象的不同，群智能算法可分为生物群体智能算法和非生物(人工)群智能算法。

生物群智能算法是一种模拟生物群体集体智慧的计算智能算法，如，细菌觅食算法(BFA)[2]模拟大肠杆菌在人体肠道内吞噬食物的行为；粒子群算法(PSO)[3]模拟鸟群寻找食物的过程；蝙蝠算法(BA)[4]模拟蝙蝠捕食的过程；蚁群算法(ACO)[5]模拟蚂蚁在寻找食物过程中的路径行为，即寻找最短路径；当然，还有萤火虫算法(FA)[6]、布谷鸟算法(CS)[7]、蛙跳算法(SFLA)[8]、鲸鱼优化算法(WOA)[9]等。

非生物群智能算法是一种模拟日常的自然现象的计算智能算法，如水滴算法(IWD)[10]，模拟水流在河道流动的过程；磁铁优化算法(MOA)[11]，模拟磁场中例子的受力过程；台风成因优化算法(TOA)[12]，模拟台风的形成过程；还包括烟花算法(Fireworks Algorithm，FWA)[13]等。

一直以来，学者们在群智能算法方面的研究热情依旧不减，但是现有研究成果及方法远远不能满足实际应用的需求。烟花算法(FWA)作为一种新型的群智能优化算法，最初为求解复合函数的全局最优化问题，具有较强的全局搜索和局部搜索能力，并且具有易实现、分布式计算机制、强鲁棒性、扩展性良好以及随机性等特点。

但在实际应用中，用标准的 FWA 算法无法有效地解决某些问题，亟待提出具有收敛速度更快、收敛精度更精确、鲁棒性更稳健的改进 FWA 算法。目前，不少研究者针对某些特定问题提出了不同的改进 FWA 算法。本文主要对烟花算法进行简要叙述，包括烟花算法提出背景、基本原理、烟花算法的改进及其应用在不同领域的优化问题，最后对烟花算法的未来研究方向进行探讨与总结。

## 2. 烟花算法

在 2010 年，Tan[13]等人提出烟花算法，该算法是模拟烟花在空中爆炸形成爆炸火花随机散落在周围，并将该过程视为粒子在解空间的寻优过程。该算法主要由爆炸强度、映射规则、选择策略、变异算子、位移操作这几个部分组成的。

### 2.1. 爆炸强度

烟花爆炸时，烟花种群的每个烟花都会生成相应的爆炸火花子种群。烟花的适应度值越好，则爆炸强度越大，即产生的爆炸火花数量越多；反之，则爆炸强度越小，即产生的爆炸火花数量越少。爆炸强度的计算如下：

$$S_i = m \frac{Y_{\max} - f(x_i) + \varepsilon}{\sum_{i=1}^{N}(Y_{\max} - f(x_i)) + \varepsilon} \quad (1)$$

式(1)中，$S_i$ 是第 $i$ 个烟花个体所产生的爆炸火花的数量；$m$ 是一个用来限制爆炸火花数量的常量；$Y_{\max}$ 是适应度值最差的烟花个体；$f(x_i)$ 是第 $i$ 个烟花个体的适应度值；$\varepsilon$ 是一个极小的常数，以避免分母为零。

为了避免爆炸火花数量过大，因此使用式(2)来进行限制。

$$\hat{S}_i = \begin{cases} round(am), S < am \\ round(bm), S > bm, a < b < 1 \\ round(S_i), others \end{cases} \quad (2)$$

其中，$round()$ 为四舍五入的取整函数；a 和 b 都是常数变量。

### 2.2. 爆炸幅度

爆炸幅度是指烟花个体发生爆炸之后所发生的位移量，其具体操作如下所示。

$$A_i = A_{\max} \frac{f(x_i) - Y_{\min} + \varepsilon}{\sum_{i=1}^{N}(f(x_i) - Y_{\min}) + \varepsilon} \quad (3)$$

$A_i$ 是第 $i$ 个烟花个体的爆炸半径，$A_{max}$ 是最大的爆炸半径，$Y_{min}$ 是最优烟花个体的适应度值。由公式 3 可知，适应度值越好的烟花个体，产生的爆炸半径越小；反之，产生的爆炸半径越大。

### 2.3. 位移操作

在烟花算法中，随机选择烟花个体的 $z$ 个维度来进行位移操作：

$$x_i^k = x_i^k + rand(0, A_i) \quad (4)$$

其中，$rand(0,A_i)$ 是一个$(0,A_i)$范围内服从均匀分布的随机函数。

### 2.4. 变异算子

随机选择烟花的 $randDim$ 个维度进行高斯变异。

用 $x_i^k$ 表示第 $i$ 个个体在第 $k$ 维的位置，那么，此时烟花的高斯变异操作为：

$$x_i^k = x_i^k * N(1,1) \quad (5)$$

其中，$N(1,1)$是服从均值为 1，方差为 1 的高斯分布函数。

### 2.5. 映射规则

在寻优的过程中，需对越界的烟花个体进行如下越界处理：

$$x_i^k = x_{\min}^k + |x_i^k| \% (x_{\max}^k - x_{\min}^k) \quad (6)$$

其中，$x_i^k$ 表示超出边界的第 $i$ 个个体在第 $k$ 维的位置；$x_{\max}^k$ 和 $x_{\min}^k$ 分别为第 $k$ 维的上边界和下边界；% 为模运算符号。

### 2.6. 选择策略

在烟花算法中，采用欧氏距离来度量任意两个个体之间的距离，即

$$R(x_i) = \sum_{j=1}^{K} d(x_i, x_j) = \sum_{j=1}^{K} \|x_i - x_j\| \quad (7)$$

其中，$d(x_i, x_j)$表示任意两个个体 $x_i$ 和 $x_j$ 之间的欧式距离；$R(x_i)$表示个体 $x_i$ 与其他个体的距离之和；$j \in K$ 表示第 $j$ 个位置属于集合 $K$；集合 $K$ 是爆炸算子和高斯变异产生的爆炸火花的位置集合。

个体选择采用轮盘赌的方式，离其他个体更远的个体具有更多的机会成为下一代烟花，每个个体被选中的概率用 $p(x_i)$ 表示，即：

$$p(x_i) = \frac{R(x_i)}{\sum_{j \in K} R(x_j)} \qquad (8)$$

综上所述，烟花算法的主要流程如算法 1 所示。

**算法 1** 烟花算法的主要流程
**Step 1** 初始化烟花种群 $x$；
**Step 2** 计算烟花适应度值；
**Step 3** 使用式(1)来计算爆炸强度；
**Step 4** 使用式(3)来计算爆炸幅度；
**Step 5** 使用式(4)来进行位移操作，并使用式（6）进行越界处理；
**Step 6** 使用式(5)产生高斯火花，并使用式（6）进行越界处理；
**Step 7** 根据 2.6 小节的选择策略来选择下一代烟花；
**Step 8** 重复 step 2~step 7，直到达到终止条件，则停止迭代。

## 3. 改进的烟花算法

自烟花算法被提出以来，学者们对烟花算法进行了不同的改进，其改进工作主要分为两类，一类是在基本烟花算法的基础上进行算子的分析和改进，另一类是与其他启发式算法的混合算法的研究[14]。

### 3.1. 算子的分析与改进

该类研究主要针对烟花算法的算子存在的缺陷，并对其进行深入的分析与研究，从而提出相应的改进策略，比如信息的有效利用、种群协同寻优、提高爆炸性能等。

Pei[15]等提出 AcFWA，将候选集的概念引入位置信息和适应度信息，来估计搜索空间的形状，生成精英解，作为一个启发式信息被引入烟花种群。

Zheng[16]等提出具有协同寻优机制的烟花算法（COFWA），将选择策略以及高斯变异进行改进，大大提高了非核心烟花的探测能力。

朱启兵[17]等提出带有引力搜索算子的烟花算法(FAGSO)，利用烟花间相互引力作用对烟花维度信息进行改善，以提高算法的优化性能。

Barraza[18]等基于模糊逻辑，对爆炸火花数目以及爆炸振幅进行调整，提高了烟花算法的性能。

Li[19]等人提出了 GFWA，引入了一种新的引导火花（GS），通过提高 FWA 中的信息利用率来进一步提高其性能。其思想是通过爆炸火花获取的目标函数信息来构造具有良好方向和自适应步长的引导向量(GV)，并将 GV 添加到烟花的位置来产生称为 GS 的精英解。

方柳平[20]等通过嵌入一种利用历史成功信息生成两种不同的学习因子来改进 dynFWA，来平衡算法的局部搜索和全局搜索能力。

Yan[21]等提出了一种具有线性降维策略的动态搜索烟花算(ld-dynFWA)，提高了算法的平衡效率和稳定性，并将其应用于求解 CNOP。

### 3.2. 混合算法的研究

该类改进烟花算法的主要思想是：将其他群智能算法的主要优点以及思想和烟花算法相融合，并提出混合的改进算法。

其中，Gao[22]等将文化算法和烟花算法混合，改变了烟花算法中爆炸火花产生方式和选择策略，提出了 CA-FWA 算法。

Zhang[23]等提出了 BBO-FWA 算法，将生物地理学优化算法的迁移算子引入烟花算法，表现出了较强勘探能力 Łapa[24]等将遗传算法与烟花算法的融合旨在解决复杂的优化问题，取得了不错的效果。

Babu[25]等人利用基于粒子群和遗传算法的烟花算法来精确识别光伏(PV)模块的双二极管模型未知参数，从而有效地解决了这一建模问题。

Bao[26]等提出了一种改进的混沌烟花算法，在求解多目标 JSP 问题上具有较高的准确性和鲁棒性。

范虹[27]等提出一种基于烟花算法的软子空间 MR 图像聚类算法，在聚类过程中引入自适应烟花算法，有效地平衡局部与全局搜索，弥补现有算法容易陷入局部最优的不足。

包晓晓[28]等通过改进烟花算法的参数提高了求解作业车间调度问题的有效性

和稳定性。Xue[29]等提出了一种自适应烟花算法（SaFWA）来优化分类模型，利用四种候选解生成策略（CSGSS）来提高解的多样性。

莫海淼[64]等将蝙蝠算法局部寻优的思想融入到烟花算法中，并且采用蝙蝠算法在全局最优附近的位置信息、蝙蝠发出的频率、全局最优以及烟花的位置信息构造了新的爆炸半径，使烟花算法在寻优的过程中能够自动地调整步长；并且使蝙蝠个体与烟花个体实现协同寻优；最后,采用"精英-随机"策略选择下一代烟花，不仅增加了烟花种群的多样性，而且减少了子代选择的时间代价。

## 4. 典型的改进烟花算法

### 4.1. 增强烟花算法

由公式(2)可知，最优烟花的爆炸半径会非常小，趋向于0。在优化过程中几乎没有进行任何搜索，实际上这样的烟花所产生的爆炸火花是几乎没有价值的，但是增加了计算时间也浪费了搜索的机会。为解决这一问题，Zheng[30]等人提出增强型烟花算法，引入了最小爆炸半径检测策略（MEACS），对最小爆炸半径做限定，计算公式如下所示：

$$A_i^k = \begin{cases} A_{min}^k, & A_i^k < A_{min}^k \\ A_i^k, & 其他 \end{cases} \quad (9)$$

其中，$A_{min}^k$ 表示第 k 维上的爆炸半径最低的检测阈值，计算公式如下：

$$A_{min}^k(t) = A_{init} - \frac{A_{init} - A_{final}}{T_{max}}\sqrt{(2T_{max}-t)t} \quad (10)$$

其中，$T_{max}$ 是最大评估次数，t 是当前迭代的评估次数，$A_{init}$ 和 $A_{final}$ 分别是初始和最终爆炸半径检测值。由于烟花的搜索过程是非线性的，本文只列出公式(10)表示非线性递减爆炸半径检测。

标准 FWA 算法的烟花所产生的火花在每一个维度的位移偏移是一样的，大大降低了种群的多样性以及勘探能力。FWA变异维度个数服从类似均匀分布，而EFWA的维度选择采用的是二项式分布方式，增加了种群的多样性。

由公式（5）可知，当随机生成的 N 值趋向于 0 时，高斯变异烟花的位置会靠近 0 且在后期很难跳出；当随机产生的 N 值比较大时，高斯变异烟花的位置超出边界，通过公式（6）可知，超出边界的烟花会映射到原点或者附近，导致了 FWA 算法在最优解在原点或者附近的测试函数上具有更好的性能，这实际上人为地加速了算法的收敛而不是算法的智能，但对其他函数而言，这就浪费了搜索资源。

为避免这些不足，新的高斯变异如公式(11)以及映射规则如公式(12)所示：

$$x_i^k = x_i^k + (x_b^k - x_i^k)*e \quad (11)$$

其中，e 为均值为 0，方差为 1 的高斯分布的随机变量；$x_b^k$ 为当前烟花种群中最优烟花在第 k 维上的位置信息。其他参数含义同式(5)。

$$x_i^k = x_{min}^k + U(0,1)*(x_{max}^k - x_{min}^k) \quad (12)$$

其中，$U(0,1)$ 是在[0,1]区间上的均匀分布随机数，其他参数含义同式(6)。

由 2.6 小节的选择策略可以看出，在进行选择下一代时需要计算任意两点的之间的欧氏距离矩阵，计算时间很大。而 EFWA 采用了"精英-随机"策略，即从烟花、爆炸火花以及高斯火花中选择最优个体，再从中随机选择其余个体。

### 4.2. 引入惯性权重的烟花算法

为加速收敛，尚菲亚[31]等将非线性惯性权重引入烟花算法(WFWA)，构造了爆炸火花新的位置更新公式如下：

$$x_j^k = wx_j^k + A_i * rand(-1,1) \quad (13)$$

$$w = (1/2)^t \quad (14)$$

其中，j=1, 2, …, $S_i$；t 是当前评估次数。在算法初期，较大的 w 有利于勘探，进行大范围搜索；随着 t 增大，w 变小有利于开采，在最优值附近进行局部搜索。

该算法与 FWA、GA、PSO 等算法进行比较，实验表明 WFWA 和 FWA 相对其他算法，在最优值在原点的测试函数上具有更快的收敛速度，且在某些函数上，WFWA 的性能优于 FWA，说明了 WFWA 的有效性；并对 WFWA 的收敛性进行了理论分析，最后得出该算法总是能收敛到全局最优。整体而言，WFWA 的性能优于

其他几种算法。

### 4.3. 动态搜索烟花算法

在 EFWA 中，最小爆炸半径检测策略过分依赖于人为设定的最大评估次数 $T_{max}$，在搜索过程中没有充分利用到搜索过程的信息。

方柳平[20]等提出一种能都动态调整核心烟花（CF）的局部和全局的搜索能力的烟花算法(dynFWA)。该算法主要包括核心烟花（core firework，CF）和非核心烟花（non-CFs）两组烟花。non-CFs 爆炸半径的计算方式与 EFWA 相同，但是没有 MEACS 检测策略；CF 爆炸半径的计算方式是根据算法在上一次迭代是否改变了最优位置的信息确定的，即当烟花种群中找到了比当前核心烟花更优的个体时，当前核心烟花将会被替代，其爆炸半径变大；反之，核心烟花爆炸半径会变小。这就是核心烟花爆炸半径的放大缩小机制。另外，对高斯变异算子进行分析，可以完全去除而不影响算法的精度，反而减少了算法的计算时间。

### 4.4. 双种群烟花算法

双种群[32]是保持种群多样性的一种有效方式。为了克服烟花算法收敛周期比较长、易陷入局部最优的问题，徐焕芬[33]等提出一种基于双种群策略的烟花算法，其主要思想是在寻优过程中，两个种群并行独立运算，交替执行爬山算子[34]和协作算子[35]。

爬山算子有利于加强在高斯变异过程中种群进行局部搜索；协作算子通过两个种群间信息共享，有利于种群进行全局搜索，维持种群多样性，避免出现"早熟"现象，这实际上也是避免爬山算子所带来的负面影响；同时算法构造了新的最大爆炸半径的计算方式，锦标赛机制的选择策略提高了算法的收敛速率。该算法与同类型的算法，如反向烟花算法[36]、混沌烟花算法[37]、遗传算法[38]和微分进化算法[39]进行比较，实验结果说明，双种群烟花算法具有更好的鲁棒性、更高的求解精度和更快的收敛速度。

### 4.5. 带有灰色关联算子的烟花算法

由公式(1)和式(3)可知，烟花适应度值越好，其烟花的爆炸半径越小，爆炸火花数越多；反之，烟花的爆炸半径就越大，爆炸火花数就越小。这种爆炸火花产生机制使得烟花之间缺乏信息交流，导致种群的多样性不高。

对于多维问题而言，寻优结果的优劣跟任意一维的值都有关系。在迭代更新过程中，烟花适应度值较差的解可能是某一维的结果较差导致的[40]。为充分利用种群之间的信息，汪菊琴等[41]提出带有灰色关联算子的烟花算法(GFA)，在寻优过程中利用灰色关联算子为每个粒子选取一个指导粒子，根据指导粒子对自身粒子维度信息进行更新，相当于在粒子周围进行了局部搜索，避免"早熟"以提高算法的寻优能力。将 GFA 与近年来提出的其他智能算法进行对比[42](DE、jDE、CLPSO、TLBO、VTTLBO),仿真结果表明：GFA 是可行的，与相关改进的烟花算法相比，GFA 整体性能最优。

## 5. 烟花算法的应用

烟花算法由于易于实现，并且还具有爆发性、多样性、简单性和随机性等特点，目前被广泛应用网络定位[43]、JSP 问题求解[28]、图像配准[44]、网络规划[45]、垃圾电子邮件检测[46]、聚类[47]、Web 服务组合优化[48]、分类[49]、关联规则[50]等多个领域。按照不同的研究对象，烟花算法的应用概况起来主要有几个方面：离散和连续优化问题，单目标和多目标优化问题以及跟其他领域的应用。本节根据不同的实际应用领域对部分案例进行叙述。

### 5.1. 离散和连续优化问题

许多实际工程与应用问题的本质是函数优化问题，即参数的设计与优化，或者将问题转化为函数优化问题求解[51]。烟花算法非常适合这类问题的求解。

宋江迪[52]以到达最优值的计算时间为度量，提出了 FWA-SFLA 混合算法，该算法对全局函数进行优化，将混合蛙跳算法的局部搜索策略引入到子代烟花的计算，使 FWA 避免过早陷入局部最优，并加快

了全局搜索能力。仿真结果表明了 FWA-SFLA 寻优具有更高效的可行性，在函数优化方面具有更好的求解精度和收敛速度。

学者将烟花算法应用到离散型优化问题。图像识别（Image Registration，IR）是指将图像进行处理后特征提取和分类，而特征选择的本质是离散空间的优化组合问题。图像识别在过去依赖于人工设置的特征，随着大数据时代的到来，对图像识别要求的准确性和时效性也越来越高。针对不同类型的图像，Bejinariu[44]等人采用了基于像素的归一化互信息配准方法。与其他算法相比，FWA 性能接近粒子群（PSO 和布谷鸟搜索（CSA）的精度。考虑计算时间，即由目标函数的评估次数所决定，FWA 比 PSO 稍慢，比 CSA 和遗传算法（GA）快得多。

Ying[53]等针对 FWA 在连续优化问题上的出色性能，提出求解组合优化问题的离散烟花算法（DFWA），应用于旅行商问题（TSP）。由于该算法的参数设置主要针对 TSP 问题的求解，还无法直接用于其他离散组合优化问题。该算法离散的烟花算法舍弃了映射规则，保留了其他部分，整体保持了烟花算法的规则。在城市规模小于 100 的 TSP 问题上，该算法能很快的就找到最优解，在城市规模约为 200 的 TSP 问题上，该算法与标准蚁群算法[5]取得相近的结果。在大规模的 TSP 问题上，该算法相对传统的蚁群算法没有优势，还需要改进。

### 5.2. 单目标和多目标优化问题

烟花算法在优化单目标离散化问题上具有较好的性能，其中 0-1 背包问题就是一个典型的离散型问题。

针对 0-1 背包问题，胡庆生[54]提出离散化烟花算法，采用了整数编码离散化机制，区别于传统 Sigmoid 函数离散化，并且引入贪心策略，对一些背包重量超出容量的不可行解和一些背包重量轻的可行解修正。用 3 个背包数据测试集对该算法进行测试，并将该算法与 DPSO、GA 和 ACO 算法进行试验对比，结果表明，离散烟花算法的具有更快的收敛速度以及更高的收敛精度。

从单目标到多目标优化问题，薛俊杰[55]等提出了一种二进制反向学习烟花算法应用于求解多维背包问题，该算法定义了用曼哈顿距离求解的二进制距离、二进制转换算子，将烟花算法的爆炸算子、变异算子离散化，并设计不完全二进制反向学习机制。用 10 个多维背包问题典型算例对该算法进行测试，并与其他智能优化算法，如二进制时变加速粒子群算法[56]、量子遗传算法[57]以及二进制布谷鸟算法[58]进行对比试验。实验结果表明，该算法在求解多维背包问题时具有不错的收敛效率、更高的寻优精度和良好的鲁棒性，尤其是在求解更高维的背包问题上，仍然具有非常不错的寻优能力。

目前，也有不少的学者将群体智能算法应用于其他多目标优化问题上。Bao[26]等人以完成时间、作业延迟时间和机器闲置时间最小为多目标的作业车间调度(Job-shop scheduling problem，JSP）优化模型，提出了一种改进的混沌烟花算法，该算法增强了局部搜索能力。实验结果表明，该算法在求解多目标 JSP 问题上具有较高的准确性和鲁棒性。后续，包晓晓[28]等人设计了四个参数实验，分析不同参数对算法求解能力的影响，找出求解 JSP 问题的较优参数，最后与 JSP 的标准问题进行了仿真对比实验。实验结果证明了烟花算法在求解 JSP 问题上具有更佳的有效性和稳定性。

此外，Yang[45]等人以满足标签 100% 覆盖率、部署更少的阅读器、使用较少的发射功率和避免信号干扰为目标的多目标 RFID（Radiofrequency Identification）网络规划优化模型，将增强烟花算法和分层方法结合，使用标准基测试集进行测试，并与 GPSO、VNPSO、GPSO-RNP、VNPSO-RNP 四种算法进行了实验对比。实验结果表明，该算法在对多目标 RFID 进行网络规划时表现更出色，可以更有效地找到最优化方案。

### 5.3. 其他应用领域

烟花算法成功地应用于开拓了新的领

域。Ding[59]等提出了基于 GPU 平台的并行烟花算法(GPU-FWA)。为了降低 FWA 中烟花之间的交互以及保持良好的并行度，提出了间隔交互式的计算爆炸半径方式，有效地提高了算法的性能。

黄伟建[60]等以任务的执行时间和虚拟机的负载均衡为多目标优化调度模型，通过在 Cloudsim 上，将烟花算法与 PSO 和 GA 进行对比，实验结果表明，烟花算法得到的实验效果更佳；另外，当种群规模不断扩大时，烟花算法的性能明显优于 PSO 算法和 GA 算法。

高文强[61]等以高效配送、降低成本、保证安全、增加收益为多目标的外卖配送路径优化模型，对烟花算法中部分参数，包括爆炸和变异算子进行了分析，有效地解决了算法中存在的不足，有效求解了基于烟花算法对外配送路径优化问题。

烟花算法在应用实际中也暴露出一些缺陷，如易出现"早熟"现象、初始点敏感，为解决这一问题，李雪源[62]等提出一种具有二进制编码机制的烟花算法(BFA)，该算法将数据聚类作为一个优化问题的解，旨在在搜索空间内寻找最优聚类中心，即最优解。实验数据表明，提出的 BFA 算法能够有效应用于数据聚类问题，且整体性能优于其他聚类算法。

## 6. 未来研究方向

烟花算法作为一种新兴的群智能优化算法，也是一个具有很大潜力的元启发式算法。截止目前，烟花算法的研究还是很初步的，在有些方面还是很肤浅甚至是空白的，迫切需要广大对此有兴趣的科研工作者继续深入研究。烟花算法的未来研究方向可归结为以下几个方面。

### 6.1. 理论研究与分析

目前，纵观 FWA 的研究成果，有不少的学者在算法的理论分析上，包括对算法的性能、收敛性、收敛速度等方面有了一定的数学分析，如在收敛性方面已经有了严格的数学证明[63]，但是总体而言烟花算法的理论研究深度远低于其他群体智能算法，还需要继续深入探讨，如，烟花算法的稳定性、参数灵敏度分析等。

### 6.2. 算子的分析与改进

尽管目前烟花算法已经有了不少的改进版本，但是都是各有各的优缺点。烟花算法包括了爆炸强度、映射规则、选择策略、变异算子、位移操作这几个部分，相对于粒子群算法而言，烟花算法的参数比较多，如何减少烟花算法的参数达到简化烟花算法的目的也是值得思考的问题。如何充分利烟花、爆炸火花、高斯火花之间的信息，建立与研究他们的协同机制和竞争机制，提高算法的求解效率。

### 6.3. 混合算法的研究

利用不同求解问题的特点设计出更高效的算法，如多目标的优化烟花算法；动态优化问题的烟花算法；并行化烟花算法等。根据现有的进化算法或者其他计算方法，如差分算法、模糊逻辑思想、聚类、分类等，尝试将他们和烟花算法结合找出更高效的方法。

### 6.4. 烟花算法应用

目前，烟花算法以及改进的烟花算法被广泛应用于多个领域。而我们应该将烟花算法以及改进的烟花算法应用到更多的领域。如，如何利用烟花算法对大数据进行处理；对智能物流配送中心选址设计更合理的优化方案等。

## 7. 结束语

烟花算法是一种新型智能优化算法，在求解优化问题过程中的表现出良好的优越性能。同时，本文介绍了烟花算法的主要思想、改进版本的烟花算法、典型的烟花算法以及烟花算法在各个领域的应用。此外，还探讨了烟花算法在未来的主要研究方向。烟花算法较其他智能算法而言，具有多点同时爆炸式搜索机制，以此平衡烟花算法的局部勘探和全局搜索能力，并且烟花算法在高维度的研究问题中表现更加优异，因此急需更多的科研工作者对烟花算法进行更加深入的研究。